\def\year{2021}\relax
\documentclass[letterpaper]{article} 
\usepackage{aaai21}  
\usepackage{times}  
\usepackage{helvet} 
\usepackage{courier}  
\usepackage[hyphens]{url}  
\usepackage{graphicx} 
\usepackage{natbib}  
\usepackage{caption} 
\usepackage{lineno}
\usepackage{adjustbox}
\usepackage{tabularx}
\usepackage{amsmath}
\usepackage{mathrsfs}
\usepackage{amssymb}
\usepackage{multirow}
\usepackage{subfigure}
\usepackage{bm}
\usepackage{times}
\usepackage{epsfig}
\usepackage{siunitx}
\usepackage{hhline}
\usepackage{booktabs}
\usepackage{csquotes}

\def\eg{\emph{e.g.}} 
\def\ie{\emph{i.e.}}

\usepackage[breaklinks=true,bookmarks=false]{hyperref}
\usepackage{hyperref}
\hypersetup{
	colorlinks=true,
	linkcolor=red,
	filecolor=blue,      
	urlcolor=black,
	citecolor=cyan,
}

\title{Decoupled and Memory-Reinforced Networks:\\
	Towards Effective Feature Learning for One-Step Person Search
}
\author{
	Chuchu Han\textsuperscript{\rm 1},
	Zhedong Zheng\textsuperscript{\rm 2,3},
	Changxin Gao\textsuperscript{\rm 1}\thanks{Corresponding author: cgao@hust.edu.cn.},
	Nong Sang\textsuperscript{\rm 1},
	Yi Yang\textsuperscript{\rm 2}
	\\
}
\affiliations{
	\textsuperscript{\rm 1} 
	Key Laboratory of Image Processing and Intelligent Control, School of Artificial Intelligence and Automation,\\ Huazhong University of Science and Technology, Wuhan, China\\
	\textsuperscript{\rm 2} ReLER, University of Technology Sydney, Australia 
	\textsuperscript{\rm 3} Baidu Research, China \\
	{\{hcc, cgao, nsang\}@hust.edu.cn}, zhedong.zheng@student.uts.edu.au, yi.yang@uts.edu.au
}

\begin{document}
	
	\maketitle
	
	\begin{abstract}
		The goal of person search is to localize and match query persons from scene images. For high efficiency, one-step methods have been developed to jointly handle the pedestrian detection and identification sub-tasks using a single network. There are two major challenges in the current one-step approaches. One is the mutual interference between the optimization objectives of multiple sub-tasks. The other is the sub-optimal identification feature learning caused by small batch size when end-to-end training. To overcome these problems, we propose a \textbf{d}ecoupled and \textbf{m}emory-\textbf{r}einforced network (DMRNet). Specifically, to reconcile the conflicts of multiple objectives, we simplify the standard tightly coupled pipelines and establish a deeply decoupled multi-task learning framework. Further, we build a memory-reinforced mechanism to boost the identification feature learning. By queuing the identification features of recently accessed instances into a memory bank, the mechanism augments the similarity pair construction for pairwise metric learning. For better encoding consistency of the stored features, a slow-moving average of the network is applied for extracting these features. In this way, the dual networks reinforce each other and converge to robust solution states. Experimentally, the proposed method obtains 93.2\% and 46.9\% mAP on CUHK-SYSU and PRW datasets, which exceeds all the existing one-step methods.
	\end{abstract}
	
	\section{Introduction}
	Person search aims at localizing and identifying a query person from a gallery of uncropped scene images. This task is generally decomposed into two sub-tasks, \ie, pedestrian detection, and person re-identification (re-ID)~\cite{zheng2019camera}. Based on this, two-step and one-step methods have been developed. Two-step methods sequentially process the sub-tasks with two separate networks, where a detector is applied to raw images for localization and a followed re-ID network extracts identification features from the detected person images~\cite{zheng2017person,lan2018person,chen2018person,han2019re,chang2018rcaa,wang2020tcts}. In contrast, one-step methods learn person localization and identification in parallel within a single network, exhibiting higher efficiency~\cite{xiao2017joint,xiao2019ian,munjal2019query,yan2019learning,dong2020instance,dong2020bi,chen2020norm,zhong2020robust}. Given an uncropped input image, one-step models predict the bounding
	boxes and the corresponding identification features of all the detected persons. 
	
	Although significant progress has been made in the one-step person search, there are two crucial issues that have not been fully solved by previous works. The first issue is that coupling the two sub-tasks in a shared network may be detrimental to the learning of each task. Specifically, popular one-step methods based on the Faster R-CNN~\cite{ren2015faster} supervise the shared Region-of-Interest (RoI) features with multi-task losses, \ie, regression loss, foreground-background classification loss, and identification loss. The competing objectives of these sub-tasks make the RoI features difficult to optimize, as pointed in~\cite{chen2018person}. The second issue lies in the constrained small batch size under the end-to-end fashion, caused by limited GPU memory. It leads to sub-optimal identification feature learning since metric learning requires vast informative similarity pairs. Previous works tackle this issue by maintaining an exponential moving average (EMA) feature proxy for every identity, \ie, a look-up table. However, when an identity is infrequently visited, its feature proxy could be outdated as the weights of the model evolve. It is unclear that this strategy could be scaled to larger datasets with numerous identities.
	
	\begin{figure}[t]
		\small
		\begin{center}
			\includegraphics[width=\linewidth]{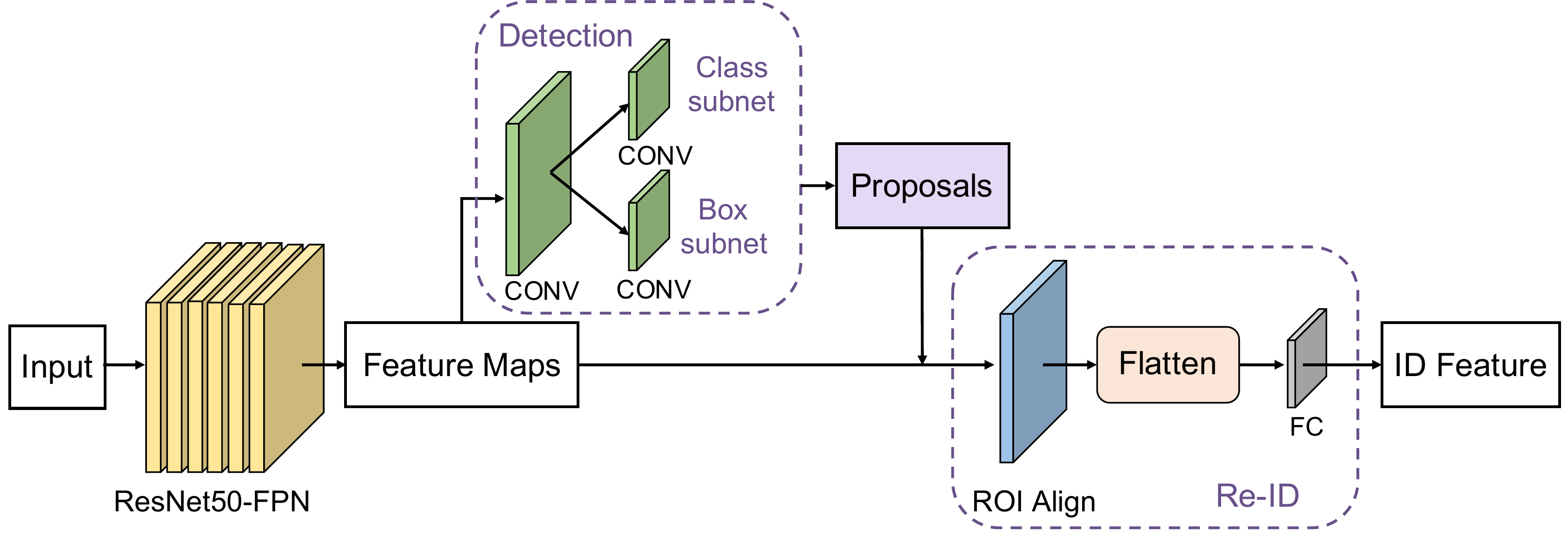}
		\end{center}
		\caption{The inference of the proposed one-step framework.}
		\label{fig:fig1}
	\end{figure}
	
	In the paper, we rethink the decoupling and integration of pedestrian detection and identification in the one-step person search framework. Considering that RoI features contain the detailed recognition patterns of detected persons, they can be specific to the re-ID task. On the other hand, bounding box regression and foreground-background classification do not have to rely on the fine-grained RoI features in light of the success of one-stage detectors. Based on these insights, we take the one-stage detector as our base network instead. As shown Fig.~\ref{fig:fig1}, foreground-background classification, regression, and re-ID subnets are branched from the layers of the feature pyramid network (FPN), which contain rich visual information and could burden multiple types of task-specific feature encoding. The fine-grained RoI features extracted from FPN are only fed into the re-ID subnet for transformation. We demonstrate that this new design makes the two sub-tasks substantially decoupled and facilitate the learning for both tasks. Specifically, the decoupled network with RetinaNet backbone~\cite{lin2017focal} achieves 6.0\% improvements on mAP compared to the popular baseline with Faster R-CNN.
	
	To further boost the identification feature learning, we build a memory-reinforced feature learning mechanism. Inspired by the recent unsupervised contrastive learning study~\cite{he2020momentum}, we memorize the feature embeddings of the recently visited instances in a queue-style memory bank for augmenting pairwise metric learning. The memorized features are consistently encoded by a slow-moving average of the network and are stored in a queue-style bank. The dual networks reinforce each other and converge to robust solution states. Experimental evidence proves that our mechanism is more effective than the look-up table.
	
	The resulting model is called decoupled and memory-reinforced network (DMRNet). Our network is easy to train because of the task decoupling in the architecture. The inference of our framework (shown in Fig.~\ref{fig:fig1}) is also very simple and efficient. In experiments, we validate the effectiveness of our methods on different one-stage detectors. Our DMRNet surpasses the state-of-the-art one-step method~\cite{chen2020norm} by 1.1\% and 2.9\% mAP on the CUHK-SYSU and PRW datasets, respectively.
	
	Our contributions can be summarized in three-folds:
	\begin{itemize}
		\item We propose a simplified one-step framework that decouples the optimization of pedestrian detection and identification. In particular, RoI features are only specific to the re-ID task, promoting the performance of both sub-tasks.
		\item We introduce a memory-reinforced mechanism for effective identification learning. A slow-moving average of the network is incorporated for consistently encoding features in a queue-style memory bank. This reinforced training makes the identification features highly discriminative.
		\item Our model is easy to train and efficient to use. It surpasses the previously best one-step methods and matches the accuracy of two-step methods.
	\end{itemize}
	
	\section{Related Work}
	\vspace{0.5em}
	\noindent\textbf{Person search.}
	Person search aims at matching a specific person among a great number of whole scene images, which has raised a lot of interest in the computer vision community recently~\cite{xiao2017joint,zheng2017person,chen2018person,lan2018person,chang2018rcaa}. In the literature, there are two approaches to deal with the problem. 
	
	\textit{Two-step methods}~\cite{zheng2017person,lan2018person,chen2018person,han2019re,chang2018rcaa,wang2020tcts} separate the person search task into two sub-tasks, the pedestrian detection, and person re-ID, trained with two independent models. Zheng~\textit{et~al.}~\cite{zheng2017person} first make a thorough evaluation on various combinations of different detectors and re-ID networks. 
	Chen~\textit{et~al.}~\cite{chen2018person} consider the contradictory objective problem existing in person search, and extract more representative features by a two-steam model. Han~\textit{et~al.}~\cite{han2019re} develop an RoI transform layer that enables gradient backpropagated from re-ID network to the detector, obtaining more reliable bounding boxes with the localization refinement. Wang~\textit{et~al.}~\cite{wang2020tcts} point out the consistency problem that the re-ID model trained with hand-drawn images are not available. They alleviate this issue by producing query-like bounding boxes as well as training with detected bounding boxes.
	
	\textit{One-step methods}~\cite{xiao2017joint,xiao2019ian,munjal2019query,yan2019learning,dong2020instance,dong2020bi,chen2020norm,zhong2020robust} develop a unified model to train the pedestrian detection and person re-ID end-to-end. Generally, this manner is more efficient with fewer parameters. 
	Xiao~\textit{et~al.}~\cite{xiao2017joint} employ the Faster R-CNN as the detector, and share base layers with the person re-ID network. Meanwhile, an Online Instance Matching (OIM) loss is proposed to enable a better convergence with large but sparse identities in the classification task. 
	To incorporate the query information into the detection network, Dong~\textit{et~al.}~\cite{dong2020bi} propose a Siamese network that both takes scene images and cropped person patches as input. With the guidance of the cropped patches, the learned model can focus more on persons. 
	As pointed out by~\cite{chen2018person}, pedestrian detection focuses on learning the commonness of all persons while person re-ID aims to distinguish the differences among multiple identities. Chen~\cite{chen2020norm} solves this problem by disintegrating the embeddings into norm and angle, which are used to measure the detection confidence and identity similarity. However, this method ignores the effect of regression loss, and excessive contexts still hamper the feature learning. Different from~\cite{chen2020norm}, we identify that the inherently defective module design is the core cause of the conflict and hinders effective feature learning.
	
	\vspace{0.5em}
	\noindent\textbf{Pedestrian detection.}
	Pedestrian Detection plays a crucial role in the person search framework. In recent years, with the advent of Convolutional Neural Network (CNN), the object detection task is soon dominated by the CNN-based detectors, which can be broadly divided into two categories: the one-stage manner~\cite{lin2017focal,redmon2016you,liu2016ssd} and two-stage manner~\cite{girshick2015fast,ren2015faster,dai2016r,he2017mask}. Due to the high efficiency, the one-stage manner has attracted much more attention recently. 
	YOLO~\cite{redmon2016you,redmon2017yolo9000} directly detects objects though a single feed-forward network with extremely fast detection speed. 
	RetinaNet~\cite{lin2017focal} solves the problem of class-imbalance by the focal loss, which focuses on learning hard examples and down-weight the numerous easy negatives. The two-stage manner is composed of a proposal generator and a region-wise prediction subnetwork ordinarily. Faster R-CNN~\cite{ren2015faster} proposes a region proposal network (RPN). It greatly reduces the amount of computation while shares the characteristics of the backbone network. Lin~\textit{et~al.}~\cite{lin2017feature} design a top-down architecture with lateral connections for building multi-level semantic feature maps at multiple scales, which is called Feature Pyramid Networks (FPN). Using FPN in a basic detection network can assist in detecting objects at different scales. Recent anchor-free detectors have raised more interest. 
	FCOS~\cite{tian2019fcos} employs the center point of objects to define positives,  then predict the four distances from positives to object boundary. 
	Reppoints~\cite{yang2019reppoints} first locate several self-learned keypoints and then predict the bound the spatial extend of objects. Without excessive hyper-parameters caused by anchors, these methods are more potential in terms of generalization ability.
	
	
	\section{Proposed Method}
	In this section, we describe the structure of the decoupled one-step person search network and present the memory-reinforced feature learning mechanism for identification. 
	
	\subsection{Decoupled one-step framework} \label{sec:3a}
\begin{figure}[t]
		\centering
		\subfigure[General one-step person search pipeline]{%
			\label{fig:diff_1}%
			\includegraphics[width=\linewidth]{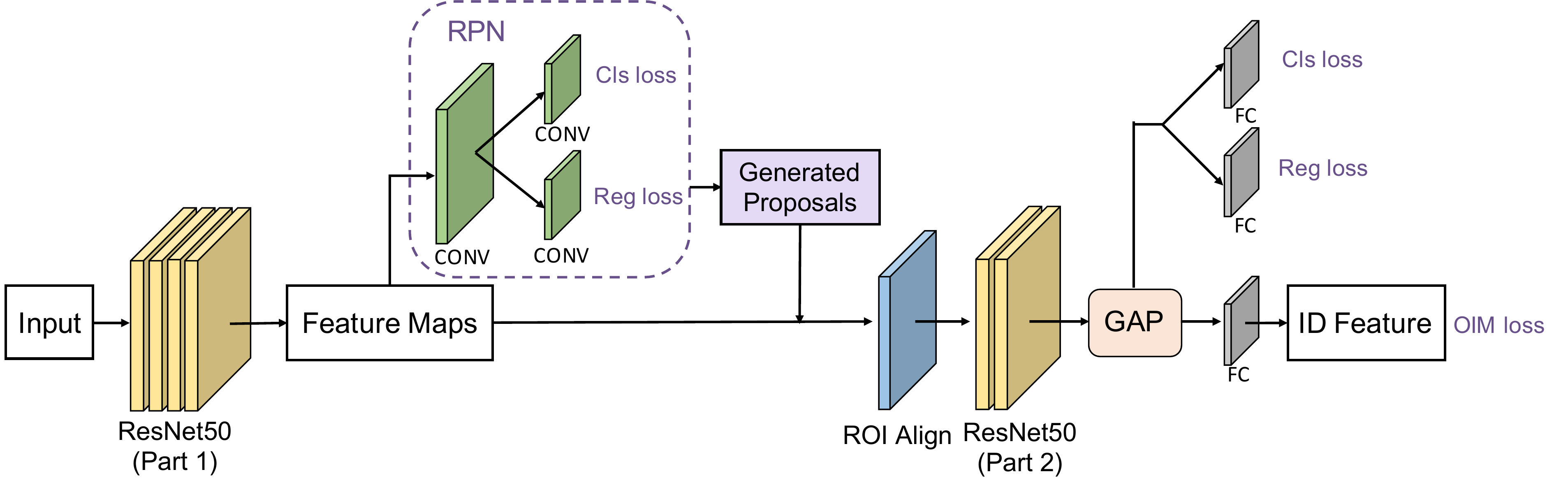}}%
		
		\subfigure[Our decoupled one-step person search pipeline]{%
			\label{fig:diff_2}%
			\includegraphics[width=\linewidth]{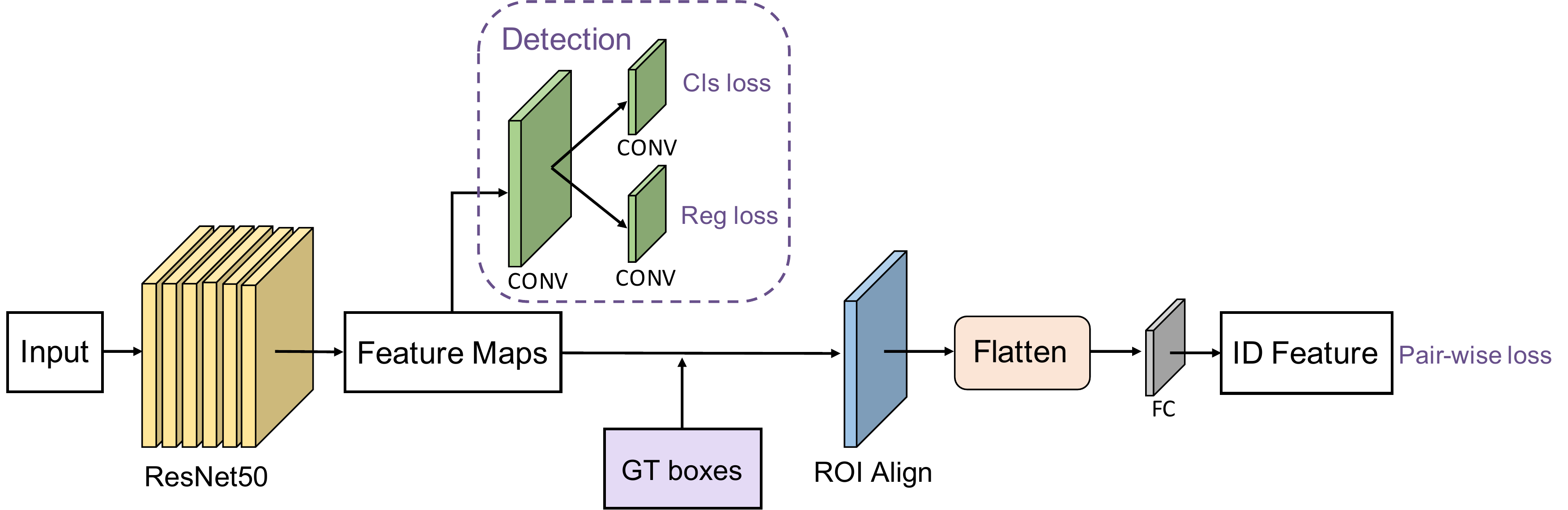}}%
		\caption{Comparisons between general training pipeline and ours. (a) General one-step person search pipeline. Multi-task losses are applied on shared RoI features. (b) Our decoupled one-step person search pipeline. The RoI features are specific to the re-ID task.}
		\label{fig:diff}
	\end{figure}

	\vspace{0.5em}
	\noindent\textbf{General one-step pipeline.}
	The first and most representative framework for one-step person search is proposed by~\cite{xiao2017joint}, and it is widely adopted in the following research work~\cite{xiao2019ian,munjal2019query,yan2019learning,dong2020instance,dong2020bi,chen2020norm,zhong2020robust}. This pipeline is based on a Faster R-CNN detector~\cite{ren2015faster}, as illustrated in Fig.~\ref{fig:diff_1}. For the re-ID module, the features are supervised by OIM loss. Together with the detection losses in RPN head and RoI head, the whole network is trained end-to-end.
	
	However, there exist contradictory objectives when supervising the shared RoI features with multi-task losses. 
	For the person search task, the detector only requires to distinguish person or background, rather than the multi-classification task in object detection. Thus, the foreground-background classification loss in the RoI head is unnecessary, even seriously affect the optimization. Evidently, foreground-background classification pursues to learn the universality of all the persons while person re-ID aims at distinguishing different persons. Moreover, the regression loss requires more information around the box boundary, while excessive contexts harm the fine-grained features for identification. 
	
	\begin{figure*}[htbp]
		\small
		\begin{center}
			\includegraphics[width=\linewidth]{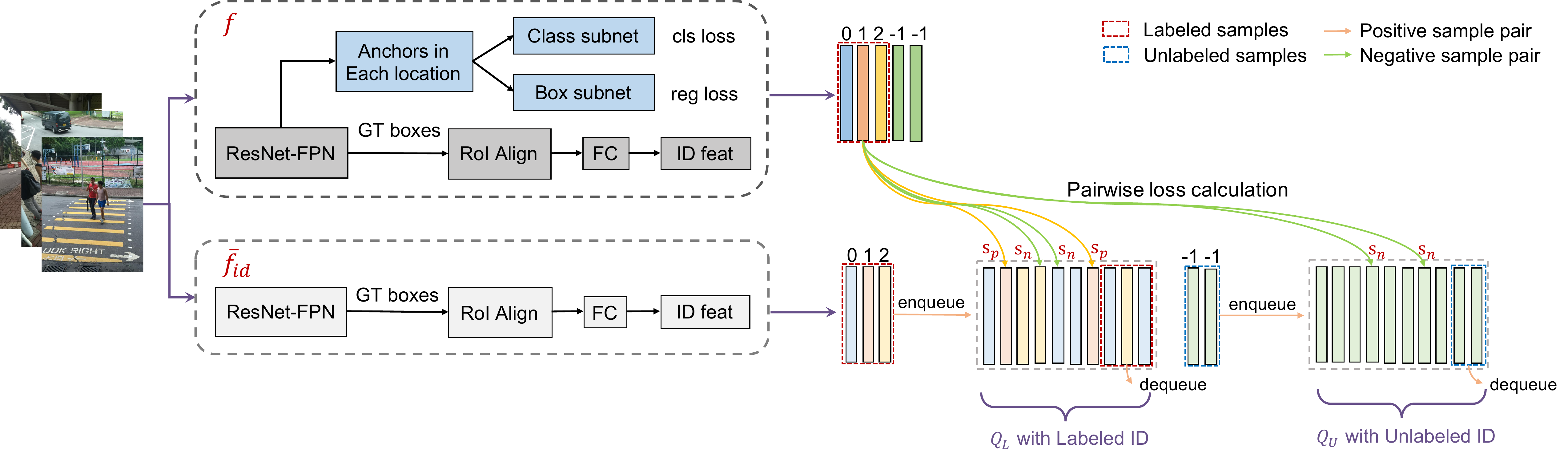}
		\end{center}
		\caption{An overview of Decoupled and Memory-Reinforced Networks (DMRNet). $\mathbf{f}$ is our decoupled person search network that trained by a SGD optimizer. $\bar{\mathbf{f}}_{id}$ is a slowly-updating network counterpart, which is utilized to consistently encode the re-ID features in the training stage. Given an input image, $\mathbf{f}$ extracts the labeled pedestrian features, termed as anchors. The features extracted by $\bar{\mathbf{f}}_{id}$ are employed to update the labeled and unlabeled queues, respectively. Thus, multiple positive and negative similarity pairs can be built between anchors and queued embeddings, supervised by a pairwise loss.}
		\label{fig:network}
	\end{figure*}
	
	\vspace{0.5em}
	\noindent\textbf{Decoupled one-step pipeline.}
	Although~\cite{chen2020norm} reconciles the conflict by factorizing embeddings into magnitude and direction for foreground scoring and re-ID, respectively, we identify that the inherently defective module design is the core cause of this issue and hinders the effective feature learning of the one-step models.
	
	In this paper, we mainly focus on learning representative RoI features for identification, instead of the multi-task losses under a shared feature space. This decoupling is based on the following considerations. First, since the RoI features contain the detailed recognition patterns of detected persons, they can be specific to the re-ID task. Second, bounding box regression and foreground-background classification do not have to rely on the fine-grained RoI features in light of the success of one-stage detectors, \eg, RetinaNet~\cite{lin2017focal}, FCOS~\cite{tian2019fcos} and Reppoint~\cite{yang2019reppoints}. 
	Based on some simplifications, we introduce the one-stage detector as our base network instead. Here we take the RetinaNet for example. As Fig.~\ref{fig:diff_2} shows, ResNet50 with a feature pyramid network (FPN) is used as the shared backbone. A class subnet and a box subnet based on FPN are employed to performs foreground-background classification and bounding box regression on each location. We add the RoI align on FPN to extract fine-grained embeddings for person re-ID. Since FPN layers include rich semantic information while RoI features contain specific content, this design makes the two sub-tasks substantially decoupled.
	Moreover, we only employ the ground truth bounding boxes to extract RoI features for re-ID training, without the usage of the predicted proposals from the regression subnet. This simplification further reduces dependencies between regression and identification. We experimentally show that using the largely reduced but accurate training bounding boxes could result in slightly better performance.
	
	\subsection{Memory-reinforced feature learning} \label{sec:3c}	
	Effective feature learning is challenging for the one-step person search. Due to the limited batch size caused by GPU memory constraints in the end-to-end fashion, it may suffer from a large variance of gradients when directly use the softmax loss or triplet loss. Previous works~\cite{xiao2017joint} use the Online Instance Matching (OIM) loss that maintains an EMA feature proxy for every identity, \ie, a look-up table. Nevertheless, due to the limited batch size, the feature proxy could be outdated as the weights of the model evolve. It is unclear that this strategy could be scaled to larger datasets with numerous identities. To keep the consistency of the comparing feature embeddings, we propose a memory-reinforced method for effective feature learning. Inspired by~\cite{he2020momentum, tarvainen2017mean}, a slowly-updating network counterpart is incorporated for yielding a consistent queue-style feature memory bank. 
	
	\vspace{0.5em}
	\noindent\textbf{Queue-style memory bank.}
	Instead of keeping the class proxy embedding within a look-up table, we maintain a queue-style memory bank. It only keeps the features of recently visited instances, avoiding features being outdated. Moreover, it decouples the memory bank size from the number of identities. This is more flexible to set the size as a hyper-parameter.
	
	\vspace{0.5em}
	\noindent\textbf{An slow-moving average of the network.} To make the stored features encoded more consistently, we introduce a slow-moving average of the network for generating features in the memory bank. We denote our decoupled network as $\mathbf{f}$, where its parameters $\theta$ are updated by the back-propagation. The slow-moving average of the network is denoted by $\bar{\mathbf{f}}_{id}$. Its parameters $\bar\theta$ are updated by EMA at each iteration:
	\begin{equation}
	\begin{aligned}
	\bar\theta \leftarrow m \bar\theta + (1-m) \theta,
	\end{aligned}
	\label{eq:encoder_momentum}
	\end{equation}
	where $m$ is the momentum factor. With a large momentum, the parameters $\bar\theta$ are updated slowly towards $\theta$, making little difference among encoders from different iterations. This ensures the consistency of the encoded features in the memory bank. Note that $\bar\theta$ is only used for extracting identification embeddings, without detection subnets. $\bar{\mathbf{f}}_{id}$ requires no gradient and brings little overhead at each iteration.
	
	\vspace{0.5em}
	\noindent\textbf{Pairwise loss for re-ID feature learning.}
	We use a pairwise loss for supervising the re-ID feature learning. The foundation of pairwise loss is to construct positive and negative pairs for metric learning.
	
	In this paper, we maintain a queue $Q_l \in \mathbb{R}^{L \times d}$ containing the features of $L$ labeled persons, and a queue $Q_u \in \mathbb{R}^{U \times d}$ containing the features of $U$ unlabeled persons, where $d$ is the feature dimension.
	Suppose the input image contains one labeled person with class-id $i$ and several unlabeled ones. The embedding of the labeled person encoded by $\mathbf{f}$ is viewed as an anchor $x_a$. The embeddings of labeled and unlabeled persons extracted by  $\bar{\mathbf{f}}_{id}$ are used to update the $Q_l$ and $Q_u$, respectively. 
	As Fig.\ref{fig:network} shows, these newest embeddings are enqueued while the outdated ones are dequeued, maintaining the queue length fixed. Assuming that there are $K$ positive samples in $Q_l$ sharing the same identity with $x_a$, and the rest $J$ ones in $Q_l$ and $Q_u$ are viewed as negative samples, the cosine similarities are denoted as $\{s_p^i\} (i=1,2,...,K)$ and $\{ s_n^j\} (j=1,2,...,J)$, respectively.
	To make every $s_p^i$ is greater than every $s_n^j$, we utilize the following loss function~\cite{sun2020circle}:
	\begin{equation}
	\begin{aligned}
	L = log[1+ \sum_{i=1}^{K} \sum_{j=1}^{J} \exp(\gamma(s_n^j-s_p^i))] \\
	\end{aligned}
	\label{eq:d}
	\end{equation}
	where $\gamma$ is a scale factor. We note that this loss formulation is the natural extension of OIM loss in the case of multiple positive similarity pairs. By the supervision of the pairwise loss, $\mathbf{f}$ and $\bar{\mathbf{f}}$ reinforce each other and their parameter spaces converge to robust solution states.
	
	\section{Experiments}
	In this section, we first describe the datasets and evaluation protocols, after which the implementation details are elaborated. Then, we conduct comprehensive ablation studies and analysis to explore the effects of different components. We further compare our method with state-of-the-art methods.
	
	\subsection{Datasets and settings} \label{sec:4a}
	\vspace{0.5em}
	\noindent\textbf{CUHK-SYSU dataset.} CUHK-SYSU~\cite{xiao2017joint} is a large scale person search dataset consisting of street/urban scenes shot by a hand-held camera and snapshots chosen from movies. There are $18,184$ images and $96,143$ annotated bounding boxes, containing $8,432$ labeled identities, and the unlabeled ones are marked as unknown instances. The training set contains $11,206$ images and $5,532$ identities, while the testing set includes $6,978$ gallery images and $2,900$ probe images. 
	
	\vspace{0.5em}
	\noindent\textbf{PRW dataset.} PRW~\cite{zheng2017person} is extracted from the video frames that are captured by six spatially disjoint cameras. There are a total of 
	$11,816$ frames with the $43,110$ annotated bounding boxes. Similar to CUHK-SYSU, it contains unlabeled identities and labeled identities ranged from $1$ to $932$. In training set, there are $5,704$ frames and $482$ identities, while the testing set includes $6,112$ gallery images and $2,057$ query images from $450$ different identities.
	
	\vspace{0.5em}
	\noindent\textbf{Evaluation protocols.} Our experiments adopt the same evaluation metrics as previous work~\cite{xiao2017joint, munjal2019query}. One is widely used in person re-ID, namely the cumulative matching cure (CMC). A matching is considered correct only if the IoU between the ground truth bounding box and the matching box is larger than 0.5. The other is the mean Average Precision (mAP) inspired by the object detection task. 
	For each query, we calculate an averaged precision (AP) by computing the area under the precision-recall curve. Then, the mAP is obtained by averaging the APs across all the queries.

	\subsection{Implementation details}  \label{sec:4c}
	For the detection network, we use the latest PyTorch implementation of RetinaNet~\cite{lin2017focal} and Reppoint~\cite{yang2019reppoints} released by OpenMMLab~\footnote{\url{https://github.com/open-mmlab/mmdetection}}~\cite{mmdetection}.
	Actually, our framework is compatible with most detectors. The queue sizes $L$ and $U$ are set to $4096$ and $4096$ for CUHK-SYSU while $1024$ and $0$ for PRW. The momentum factor $m$ is set to $0.999$, and the scale factor $\gamma$ is set to $16$.
	The batch size is $3$ due to the limitation of GPU memory. We use the batched Stochastic Gradient Descent (SGD) optimizer with a momentum of $0.9$. The weight decay factor for L2 regularization is set to $5\times10^{-4}$. As for the learning rate strategy, we use a step decay learning rate schedule with a warm-up strategy, and our model is trained for $12$ epochs totally. The base learning rate is $0$, which warms up to $1\times10^{-3}$ in the first $500$ iterations, then decays to $1\times10^{-4}$ and $1\times10^{-5}$ after $8$ and $11$ epochs. 
	All experiments are implemented on the PyTorch framework, and the network is trained on an NVIDIA GeForce GTX 1080 Ti. We also use PaddlePaddle to implement our method and achieve similar performance.
	
	
	\subsection{Ablation study}  \label{sec:4d}
	In this section, we conduct detailed ablation studies to evaluate the effectiveness of each component. First, we explore the effect of different network designs. Second, we analyze two loss mechanisms under different sizes of memory banks. Third, we exhibit the performance of our proposed method under different settings. 
	
	\begin{figure}[htbp]
		\small
		\begin{center}
			\includegraphics[width=\linewidth]{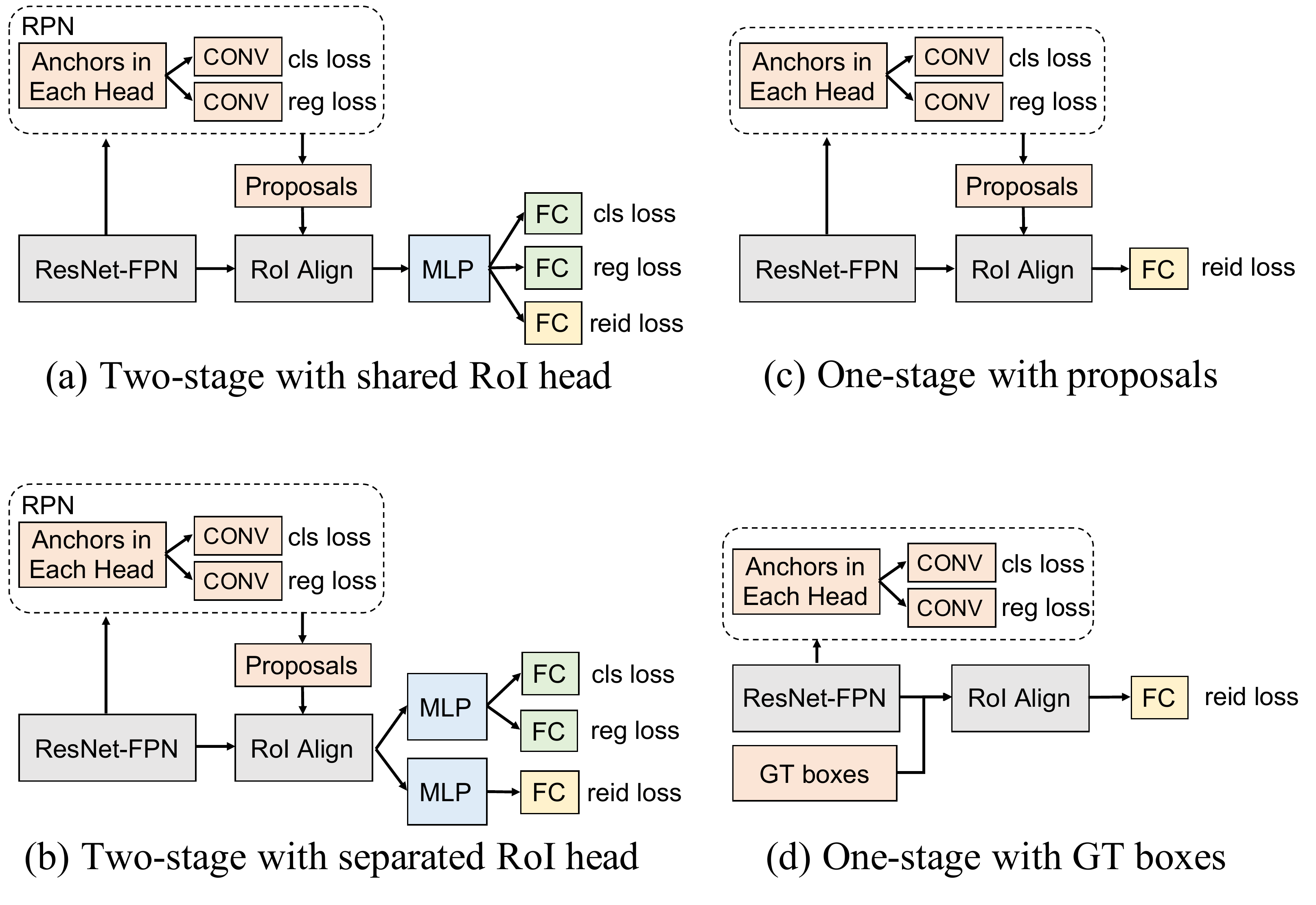}
		\end{center}
		\caption{Comparisons on different network designs. (a) is the general person search pipeline with a shared RoI head. (b) eases the coupling with separated RoI heads for detection and re-ID. (c) discards the detection losses in RoI head and the RoI features are specific for identification. (d) removes the selected proposals and only uses the GT boxes.}
		\label{fig:ablation_network}
	\end{figure}
	
	\vspace{0.5em}
	\noindent\textbf{Why is one-step inferior than two-step?} 
	To investigate what causes the poor performance in one-step person search, we conduct several experiments to illustrate the comparisons among different network options, as shown in Fig.~\ref{fig:ablation_network}. Detailed results are shown in Tab.~\ref{fig:ablation_network}. 
	
	For fair comparisons, we incorporate FPN into the general one-step framework~\cite{xiao2017joint} as our baseline (a), and this improves the performance by a large margin. When it comes to tangled sub-tasks (detection and re-ID) conflict in the one-step person search, it is natural to think about decoupling different tasks from the backbone. (b) employs separated RoI heads for detection and re-ID training. In Tab.~\ref{fig:ablation_network}, the results perform better than a shared RoI head manner on both re-ID and detection tasks. This indicates the severe coupling network harms the optimization on both sub-tasks when sharing feature space, and it can be mitigated with a simple head disentanglement. 
	
	To further eliminate the conflict, we only focus on identification feature learning instead of the multi-task loss under the shared RoI features. As shown in (c), a one-stage detector can be well incorporated and the RoI features are specific for identification. This manner surpasses (b) on both re-ID and detection performance. It shows the decoupling benefits the optimization on two sub-tasks. Note that the performance of separated trained detectors for one-stage (RetinaNet) or two-stage (Faster RCNN) is almost the same. 
	
	In (a)-(c), except for the ground truth boxes, the selected proposals (IoU$>$0.5) are also used to extract features for re-ID training. We further simplify the network by using only ground truth bounding boxes. Although the improvement is marginal, it saves much computational cost in training. Finally, based on our proposed memory-reinforced feature learning, the performance achieves 91.2\%/92.5\% on mAP/rank-1 on the CUHK-SYSU dateset.
	\begin{table}[htbp]
		\footnotesize
		\begin{center}
			\caption{Comparisons of different network designs on the CUHK-SYSU dataset. The performance of re-ID and detector trained in a single network is represented. Detector-S denotes the result of the separated trained detector.}
			\label{tab:ablation_network}
			\vspace{1mm}
			\begin{adjustbox}{max width=0.46\textwidth}
				\begin{tabular}{c|cc|c|c}
					\toprule
					\multirow{2}*{Methods} &\multicolumn{2}{c}{Re-ID}&Detector&Detector-S\\
					~&mAP&Rank-1&mAP&mAP\\
					\midrule			
					\midrule
					Faster+OIM\cite{xiao2017joint} & 75.5 & 78.7 & -& - \\	
					\midrule
					Faster(FPN) + OIM w/ (a)  & 84.3 & 84.6&86.9 &\multirow{2}*{92.2}\\
					Faster(FPN) + OIM w/ (b) & 87.5 & 87.7 &89.8&~\\
					\midrule
					Retina(FPN) + OIM w/ (c) & 90.0 & 90.8 &91.2& \multirow{3}*{92.3}   \\
					Retina(FPN) + OIM w/ (d) & 90.3 & 91.0 &91.4& ~  \\
					Retina(FPN) + DMR w/ (d) & 91.2 & 92.5 &91.3& ~  \\
					\bottomrule
				\end{tabular}
			\end{adjustbox}
		\end{center}
	\end{table}

	\vspace{0.5em}
	\noindent\textbf{Effectiveness on different detectors.} 
	In order to evaluate the expandability of our method, we incorporate different detection networks into our framework, including RetinaNet~\cite{lin2017focal} and Reppoint~\cite{yang2019reppoints}.
	The separated trained detectors reach 92.3\% and 93.1\% on mAP, respectively. We show the person search results in Tab.~\ref{tab:ablation_detector} under different settings. When only perform the decoupled network, the results have achieved 90.3\% and 92.4\% rank-1 with RetinaNet and Reppoint, respectively. The performance is further promoted when employing the memory-reinforced method for training. This confirms the effectiveness and robustness of our method when extended to different detectors.
	Moreover, we show the experimental results under different resolutions. It is obvious that a larger image reaches higher performance. 
	
	\begin{table}[htbp]
		\footnotesize
		\begin{center}
			\caption{The results on the CUHK-SYSU and PRW datasets with different detectors. D denotes the decoupled framework while DMR means our decouple and memory-reinforced network.}
			\label{tab:ablation_detector}
			\vspace{1mm}
			\begin{adjustbox}{max width=0.48\textwidth}
				\begin{tabular}{c|c|cc|cc}
					\toprule
					\multirow{2}*{Methods} &\multirow{2}*{Resolution} & \multicolumn{2}{c|}{CUHK-SYSU} & \multicolumn{2}{c}{PRW} \\
					~&~&mAP&Rank-1&mAP&Rank-1\\
					\midrule			
					\midrule
					Retina+D&1333*800& 90.3 & 91.0 & 36.1 & 73.6\\
					Retina+DMR&1333*800& 91.2 & 92.5 & 44.6 & 82.0\\
					Retina+DMR&1500*900& 91.6 & 93.0 & 46.1 & 83.2 \\
					\midrule
					Reppoint+D&1333*800& 92.4 & 93.2 & 39.1&73.6\\
					Reppoint+DMR&1333*800& 92.9 & 93.7 &46.0 & 83.2\\
					Reppoint+DMR&1500*900& 93.2 & 94.2 & 46.9 & 83.4\\
					\bottomrule
				\end{tabular}
			\end{adjustbox}
		\end{center}
	\end{table}
	
\begin{figure}[t]
		\small
		\begin{center}
			\includegraphics[width=\linewidth]{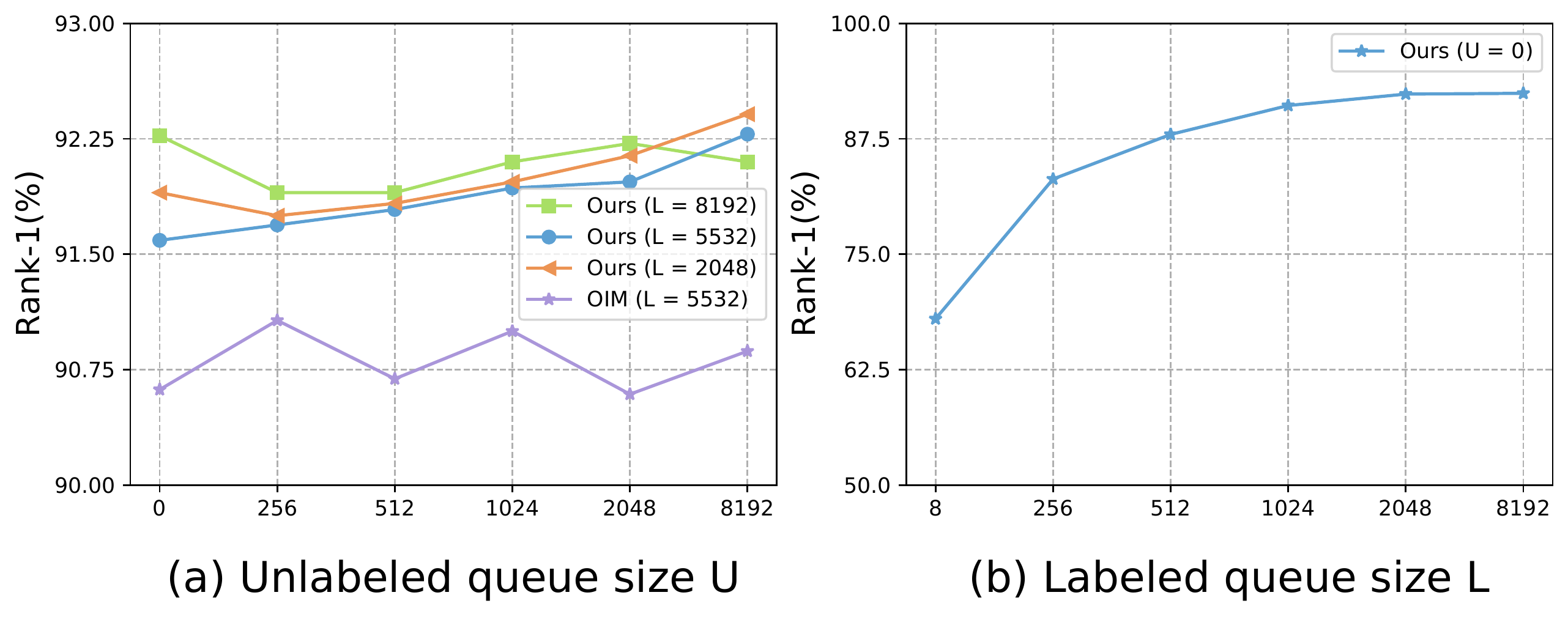}
			\caption{Comparison between OIM loss and our DMRNet with different sizes of memory bank. The numbers of labeled and unlabeled samples are denoted as $L$ and $U$, respectively.}
			\label{fig:ablation_loss}
		\end{center}
	\end{figure}

	\vspace{0.5em}
	\noindent\textbf{Different sizes of the memory bank.} 
	We analyze the effect of different memory bank sizes on two metric learning mechanisms, OIM loss and our  memory-reinforced mechanism. They are implemented on the same network, as described in Fig.~\ref{fig:ablation_network}(d). $L$ is the length of the look-up table or queue with labeled samples, and $U$ is the length of the queue with unlabeled ones. The comparisons are shown in Fig.~\ref{fig:ablation_loss}, from which we have the following observations. 
	\begin{itemize}
		\item[-] To explore the effect of unlabeled samples, we compare OIM (L=5532) with our method ($L$=2048/5532/8192) under different sizes of $U$. As shown in Fig.~\ref{fig:ablation_loss} (a), the performance of our method is constantly promoted as $U$ increases when $L$=2048/5532. This shows that exploring more negative samples is better for optimization. The relatively large size of the labeled queue ($L=8192$) cannot benefit from $U$. This is reasonable as a larger $L$ has provided sufficient negative samples. For OIM loss, there is no significant improvement when $U$ increases. Due to the lack of feature consistency, more sample pairs contribute little to the result.
		\vspace{1mm}
		\item[-] As Fig.~\ref{fig:ablation_loss}(a) shows, when $U$ is set to zero, our method benefits from a larger $L$. This is intuitive since more positive/negative sample pairs can be exploited.
		\vspace{1mm}
		\item[-] From Fig.~\ref{fig:ablation_loss}(a)(b), it can be observed that when the two methods reach the same performance, our method is more efficient (L=2000, U=0) than OIM (L=5532, U=5000).
	\end{itemize}
	
	\vspace{0.5em}
	\noindent\textbf{Momentum factor.} The performance of our method with different momentum factors is shown in Tab.~\ref{tab:ablation_m}. We obtain the optimal result when $m$ is set to $0.999$. This indicates a relatively large momentum facilitates learning discriminative identification features. When $m$ is zero, it means the parameters of $\mathbf f$ and $\bar{\mathbf f_{id}}$ are identical. Surprisingly, with the least consistent encoding, our mechanism still slightly outperforms the look-up table by $0.3\%$ mAP and $0.6\%$ rank-1, showing the effectiveness of the queues.
	\begin{table}[htbp]
		\footnotesize
		\begin{center}
			\caption{The results with different momentum factors $m$ on the CUHK-SYSU dataset.}
			\label{tab:ablation_m}
			\vspace{1mm}
			\begin{adjustbox}{max width=0.48\textwidth}
				\begin{tabular}{c|cccccc}
					\toprule
					$m$&0&0.5&0.9&0.99&0.999&0.9999\\
					\midrule			
					\midrule
					Rank-1&91.6& 91.6& 91.7&91.7&92.5& 91.6\\
					mAP&90.6&90.7&90.9&90.9&91.2&90.4\\
					\bottomrule
				\end{tabular}
			\end{adjustbox}
		\end{center}
	\end{table}

	\subsection{Comparisons with the state-of-the-art methods} \label{sec:4e}
	
	In this section, we compare our proposed DMRNet with current state-of-the-art methods on person search in Tab.~\ref{tab:sysu&prw}.
	The results of two-step methods~\cite{chang2018rcaa,chen2018person,lan2018person,han2019re,wang2020tcts} are shown in the upper block while the one-step methods~\cite{xiao2017joint,xiao2019ian,liu2017neural,yan2019learning,zhang2020tasks,munjal2019query,chen2020norm} in the lower block.
	
	\vspace{0.5em}
	\noindent\textbf{Evaluation On CUHK-SYSU.} 
	The performance comparison between our network and existing competitive methods on the CUHK-SYSU dataset is shown in Tab.~\ref{tab:sysu&prw}. When the gallery size is set to 100, our proposed DMRNet reaches 93.2\%mAP and 94.2\%rank-1. It can be seen that our method significantly outperforms all other one-step methods, as well as most two-step ones. 
	
\begin{figure}[htbp]
		\begin{center}
			\includegraphics[width=\linewidth]{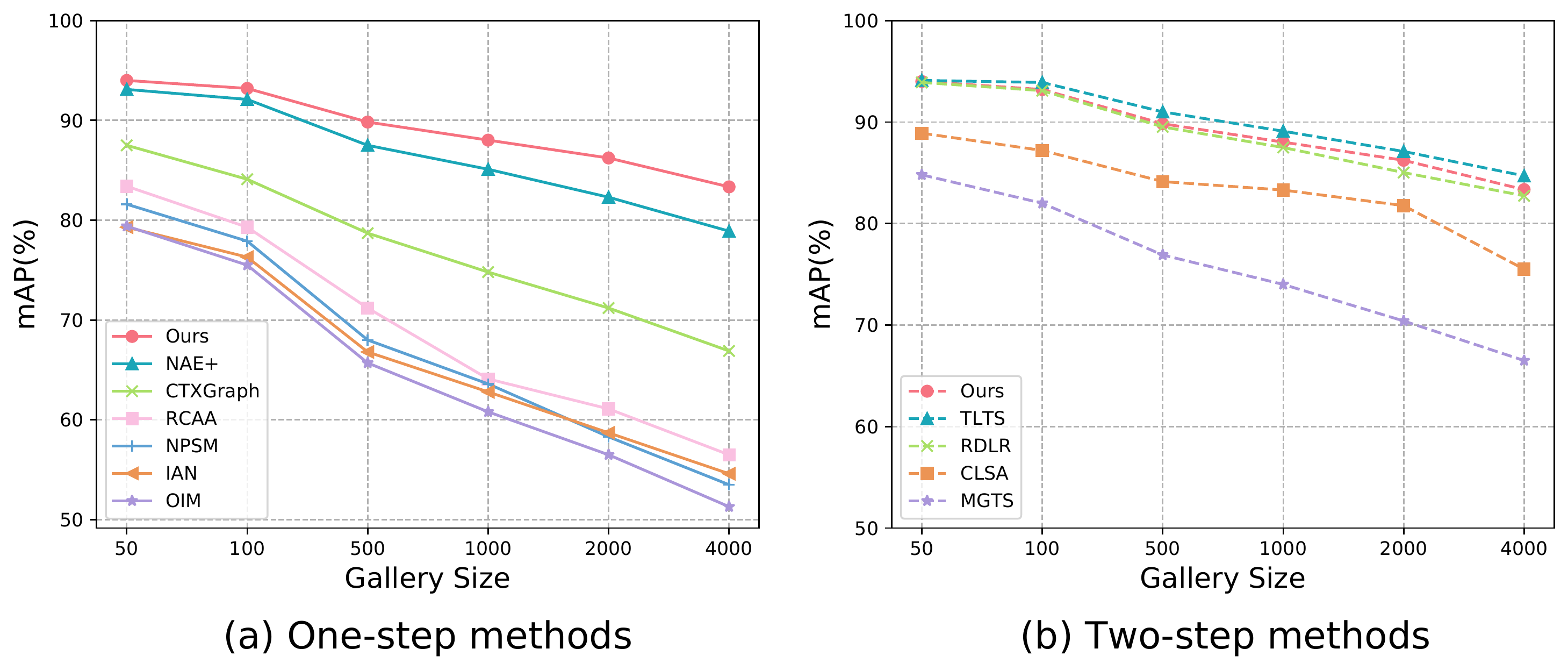}
		\end{center}
		\caption{Comparisons with different gallery sizes on the CUHK-SYSU dataset. (a) and (b) shows the comparisons with one-step methods and two-step methods, respectively.}
		\label{fig:gallery_size}
	\end{figure}

	To evaluate the performance consistency, we also compare with other competitive methods under varying gallery sizes of $[50, 100, 500, 1000, 2000, 4000]$. Fig.~\ref{fig:gallery_size} (a) shows the comparisons with one-step methods while (b) with two-step ones. 
	It can be seen that the performance of all methods decreases as the gallery size increases. This indicates it is challenging when more distracting people are involved in the identity matching process, which is close to real-world applications. 
	Our method outperforms all the one-step methods while achieving comparable performance to the two-step methods under different gallery sizes.

	\vspace{0.5em}
	\noindent\textbf{Evaluation On PRW.} 
	We further evaluate our method with the competitive techniques on the PRW dataset, shown in Tab.~\ref{tab:sysu&prw}. We follow the benchmarking setting~\cite{zheng2017person} that the gallery contains all the 6112 testing images. Compare with the current state-of-the-art one-step method~\cite{chen2020norm}, it can be seen that our method outperforms it by 2.9\%/2.2\% on mAP and rank-1. Moreover, the mAP even surpasses the best two-step method~\cite{wang2020tcts} by a marginal improvement.
	\begin{table}[t]
		\footnotesize
		\begin{center}
			\caption{Experimental comparisons with state-of-the-art methods on the CUHK-SYSU and PRW dataset.}
			\label{tab:sysu&prw}
			\vspace{1mm}
			\begin{adjustbox}{max width=0.48\textwidth}
				\begin{tabular}{l|cc|cc}
					\toprule
					\multirow{2}*{Methods} & \multicolumn{2}{c|}{CUHK-SYSU} & \multicolumn{2}{c}{PRW} \\
					~&mAP&Rank-1&mAP&Rank-1\\
					\midrule			
					\midrule
					\multicolumn{5}{l}{\textbf{\textit{Two-Step Methods}}}  \\
					RCAA~\cite{chang2018rcaa} & 79.3 & 81.3 &-&- \\
					MGTS~\cite{chen2018person} & 83.0 & 83.7& 32.6 & 72.1 \\
					CLSA~\cite{lan2018person} & 87.2 & 88.5& 38.7  & 65.0  \\
					RDLR~\cite{han2019re} & 93.0 & 94.2 & 42.9 & 70.2  \\
					TCTS~\cite{wang2020tcts} & \textbf{93.9} & \textbf{95.1}&\textbf{ 46.8} & \textbf{87.5}  \\
					\midrule
					
					\multicolumn{5}{l}{\textbf{\textit{One-Step Methods}}}  \\
					OIM~\cite{xiao2017joint} & 75.5  & 78.7& 21.3& 49.9  \\
					IAN~\cite{xiao2019ian} & 76.3& 80.1& 23.0 & 61.9  \\
					NPSM~\cite{liu2017neural}  & 77.9& 81.2& 24.2& 53.1   \\
					CTXGraph~\cite{yan2019learning} & 84.1& 86.5 & 33.4 & 73.6  \\
					DC-I-Net~\cite{zhang2020tasks} & 86.2& 86.5 & 31.8 & 55.1  \\
					QEEPS~\cite{munjal2019query} & 88.9& 89.1& 37.1 & 76.7  \\
					NAE~\cite{chen2020norm} & 91.5& 92.4 & 43.3 & 80.9 \\
					NAE+~\cite{chen2020norm} & 92.1& 92.9 & 44.0 & 81.1 \\
					Ours &\textbf{93.2} & \textbf{94.2} &\textbf{46.9}&\textbf{83.3} \\
					\bottomrule
				\end{tabular}
			\end{adjustbox}
		\end{center}
	\end{table}
	
	\vspace{0.5em}
	\noindent\textbf{Runtime Comparisons.} 
	To compare the efficiency of our framework with other methods in the inference stage, we report the average runtime of the detection and re-ID for a panorama image. For a fair comparison, we test the models with an input image size as $900 \times 1500$, which is the same as other works~\cite{chen2020norm,munjal2019query,chen2018person}. Since the methods are implemented with different GPUs, we also report the TFLOPs. As shown in Tab.~\ref{tab:runtime}, upon normalization with TFLOPs, our framework is 5.73 times faster than the two-step method MGTS~\cite{chen2018person}. Moreover, our method is more efficient than NAE+~\cite{chen2020norm}, which is the current state-of-the-art one-step method.
	\begin{table}[htbp]
		\footnotesize
		\begin{center}
			\caption{Runtime comparisons of different methods.}
			\label{tab:runtime}
			\vspace{1mm}
				\begin{tabular}{c|ccc}
					\toprule
					Methods& GPU & TFLOPs& Time  \\
					\midrule
					\midrule
					MGTS~\cite{chen2018person}&K80 &8.7&1296 \\
					QEEPS~\cite{munjal2019query}&P6000 & 12.0&300 \\
					NAE~\cite{chen2020norm}&V100 &14.1&83 \\
					NAE+~\cite{chen2020norm}&V100& 14.1&98 \\
					Ours&V100 &14.1&66  \\
					\bottomrule
				\end{tabular}
		\end{center}
	\end{table}
	
	\section{Conclusion}
	In this work, we propose a novel one-step method for person search, called the decoupled and memory-reinforced network. Extend from the one-stage detector, our multi-task learning framework substantially decouples the two sub-tasks. The RoI features are specific to identification, rather than supervised by multi-task losses. It also incorporates a slow-moving average of the network for yielding a consistently encoded queue-style feature memory bank. By mining informative features, our model could learn highly discriminative identification feature embeddings. Due to the massive simplification of the pipeline design, our model is easy to train and efficient to use. It sets a new state-of-the-art among one-step methods and outperforms a lot of existing two-step methods. We believe that our findings can encourage a shift in the framework of the one-step person search and drive more research on this field.
	\section{ Acknowledgments}
	This work was supported by the Project of the National Natural Science Foundation of China No. 61876210, the Fundamental Research Funds for the Central Universities No.2019kfyXKJC024, and the 111 Project on Computational Intelligence and Intelligent Control under Grant B18024.
	
	{\small
		\bibliographystyle{aaai21}
		\bibliography{ref}
	}
	
\end{document}